\documentclass[10pt,twocolumn,letterpaper]{article}

\usepackage{cvpr}              

\usepackage{amsfonts}       
\usepackage{nicefrac}       
\usepackage{microtype}      
\usepackage{xcolor}         
\usepackage{graphicx}
\usepackage{amssymb, amsfonts}
\usepackage{amsmath}
\usepackage{multirow}
\usepackage{colortbl}
\usepackage{xcolor}
\usepackage{tabularx}

\definecolor{rank1}{RGB}{226, 164, 145} 
\definecolor{rank2}{RGB}{235, 197, 185} 
\definecolor{rank3}{RGB}{244, 227, 222}

\definecolor{cvprblue}{rgb}{0.21,0.49,0.74}
\usepackage[pagebackref,breaklinks,colorlinks,allcolors=cvprblue]{hyperref}

\def\paperID{19240} 
\def\confName{CVPR}
\def\confYear{2026}

\title{Enhancing Video Inpainting with Aligned Frame Interval Guidance}

\author{
    Ming Xie\textsuperscript{1,2}\thanks{Equal contribution.} \quad
    Junqiu Yu\textsuperscript{1}\footnotemark[1] \quad 
    Qiaole Dong\textsuperscript{1} \quad
    Xiangyang Xue\textsuperscript{1} \quad
    Yanwei Fu\textsuperscript{1,2}\thanks{Corresponding author.}
    \\
    \textsuperscript{1}Fudan University \quad
    \textsuperscript{2}Shanghai Innovation Institute
    \\
    {\tt\small mxie24@m.fudan.edu.cn, yanweifu@fudan.edu.cn}
}

\begin{document}
\maketitle
\begin{abstract}
Recent image-to-video (I2V) based video inpainting methods have made significant strides by leveraging single-image priors and modeling temporal consistency across masked frames. Nevertheless, these methods suffer from severe content degradation within video chunks. Furthermore, the absence of a robust frame alignment scheme compromises intra-chunk and inter-chunk spatiotemporal stability, resulting in insufficient control over the entire video. To address these limitations, we propose VidPivot, a novel framework that decouples video inpainting into two sub-tasks: multi-frame consistent image inpainting and masked area motion propagation. Our approach introduces frame interval priors as spatiotemporal cues to guide the inpainting process. To enhance cross-frame coherence, we design a FrameProp Module that implements a frame content propagation strategy, diffusing reference frame content into subsequent frames via a splicing mechanism. Additionally, a dedicated context controller encodes these coherent frame priors into the I2V generative backbone, effectively serving as soft constrain to suppress content distortion during generation. Extensive evaluations demonstrate that VidPivot achieves competitive performance across diverse benchmarks and generalizes well to different video inpainting scenarios. 
\end{abstract}    
\section{Introduction}
\begin{figure}
\begin{center}
\includegraphics[width=\linewidth]{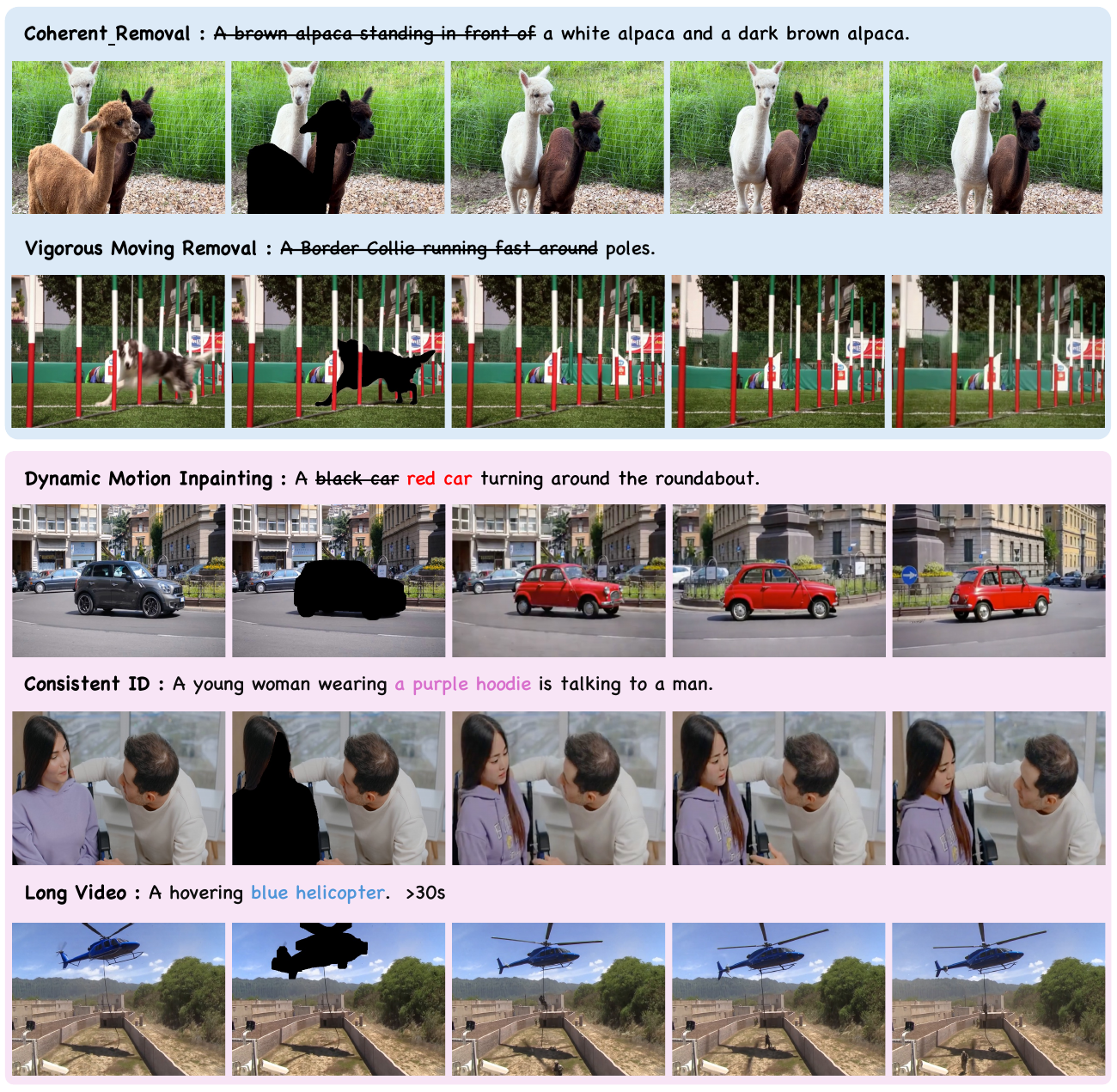}
\end{center}
\vspace{-6pt}
\caption{The inpainting results of our VidPivot. VidPivot can preserve consistent ID across inpainted videos, achieve coherent and stable masked area generation.\label{image:teaser}}
\end{figure}
Video inpainting, the task of generating or removing specific elements within a video based on given masks, sits at the intersection of creativity and perception, representing one of the most challenging problems in computer vision. Its inherent difficulty stems from the dual requirement of spatial fidelity and temporal coherence: each inpainted frame must be visually indistinguishable from its surroundings, while the sequence as a whole must evolve naturally over time, free from temporal artifacts, distortion, or flickering. Recent advances in Diffusion Transformers (DiTs)~\cite{ho2020denoising,peebles2023scalable,rombach2022high} and video generative priors~\cite{hong2022cogvideo,yang2024cogvideox,wang2025wan} have brought unprecedented generative quality, offering a promising avenue to tackle temporal consistency with remarkable fidelity.

Traditional approaches, relying predominantly on 3D CNNs~\cite{chang2019free,hu2020proposal,wang2019video}, often struggle when capturing long-range temporal dynamics, producing temporally unstable results. Subsequent methods enhanced temporal modeling via optical flow guidance~\cite{zhou2023propainter} or by leveraging pretrained image-to-video (I2V) generative models~\cite{zi2025cococo,wan2024unipaint,li2025diffueraser,gu2024advanced,bian2025videopainter, jiang2025vace}. For instance, COCOCO~\cite{zi2025cococo} injects motion-aware modules into T2I backbones for text-guided video inpainting, while DiffuEraser~\cite{li2025diffueraser} fuses BrushNet~\cite{ju2024brushnet} with diffusion models to achieve consistency over time. Complementarily, video editing methods~\cite{meng2021sdedit,ceylan2023pix2video,geyer2023tokenflow} employ attention or structure guidance to manipulate appearance, but they fall short in reconstructing missing semantics, particularly when identity alignment and temporal coherence are critical.

In some real-world scenarios, video inpainting necessitates processing videos in chunks. Image-to-Video based models are inherently well-suited to address this challenge by employing a sliding window, where the last frame of the preceding chunk serves as the initial frame for the subsequent chunk to maintain inter-chunk consistency. Despite these successes, existing methods often rely solely on a single inpainted frame~\cite{rombach2022high,fluxfill2024} as reference and delegate whole motion learning to a pretrained image-to-video model~\cite{guo2023animatediff,hong2022cogvideo,yang2024cogvideox}. This implicit handling of masked area content can lead to severe artifacts or content degradation when motion estimation is suboptimal, especially in the case of dynamic scenes, resulting in incomplete regions, content distortion, or a failure to maintain spatiotemporal consistency, as illustrated in Fig.~\ref{image:inpaintingcompare} and Fig.~\ref{image:ablation_vis}. 

To address these challenges, we propose VidPivot, a novel video inpainting framework that systematically decouples the task into two complementary subproblems: maintaining per-frame spatial coherence and ensuring cross-frame temporal consistency. Specifically, we introduce the FrameProp module for multi-frame consistent image inpainting, which ensures temporal coherence across the video while preserving spatial consistency within each frame. To further augment temporal stability, we propose a first-frame propagation mechanism. This mechanism diffuses the content from the initial inpainted frame to subsequent sampled frames through a learnable splicing operation. Complementing this, we introduce masked area content propagation to preserve short-range temporal and motion dynamics. Central to the VidPivot framework is the strategic use of spaced-frame guidance. By leveraging sparsely sampled frames, our model is supplied with structured temporal cues that guide the inpainting process. Moreover, we incorporate a context controller module that encodes and injects contextual video information, capturing coarse camera and motion dynamics while constraining the inpainting process across the entire sequence. Qualitative results on real-world videos demonstrate that VidPivot preserves consistent identities across inpainted and edited sequences, achieves stable motion generation and viewpoint transformations, and generalizes well to a varity of scenarios, as illustrated in Fig.~\ref{image:teaser}. 

Our contributions can be summarized as follows:
\begin{itemize}
    \item We propose a novel framework VidPivot for video inpainting, which successfully maintains spatiotemporal consistency and generalizes well to various video scenarios.
    \item We propose the FrameProp Module, an innovative component that generates consistent context video streams using priors from image inpainting model.
    \item We develop a context controller that effectively extracts and injects contextual features from context video into a diffusion process of whole video inpainting.
    \item Extensive experiments demonstrate that VidPivot achieves state-of-the-art performance in different benchmarks. It exhibits strong foreground alignment and motion stability.
\end{itemize}

\section{Related Works}
\textbf{Video Inpainting.} Video inpainting aims to model the spatiotemporal dependencies within masked regions and has seen significant progress in recent years. Early approaches~\cite{chang2019free,hu2020proposal,wang2019video,chang2019learnable} typically employed 3D CNNs to capture local spatiotemporal patterns. However, they always struggled with long-range temporal propagation. 
More recent methods~\cite{zi2025cococo,wan2024unipaint,li2025diffueraser,gu2024advanced,bian2025videopainter,cho2025elevating,jiang2025vace} have shifted toward leveraging pretrained video generative models, which enable temporally coherent inpainting by better capturing motion dynamics and long-range temporal consistency. Among them, the most similar work to ours is VideoPainter~\cite{bian2025videopainter} and VACE~\cite{jiang2025vace}, both leverage video diffusion models for temporal modeling and VideoPainter use an ID adapter for ID preservation. However,VideoPainter struggles to preserve consistent ID in some inpainting scenarios and introduces noticeable motion artifacts due to error accumulation of ID clips. In contrast, our VidPivot employs FrameProp to generate consistent frames for direct guidance and enjoys persistent ID alignment.

\noindent\textbf{Video Editing.} Video editing methods~\cite{ku2024anyv2v,cheng2023consistent,esser2023structure,wang2023videocomposer,jiang2025vace} primarily focus on modifying the properties of existing elements, such as their color, style, or texture. Some works~\cite{cheng2023consistent,singer2024video} perform video editing through text-based control, while others incorporate structured data~\cite{esser2023structure,wang2023videocomposer,liang2024flowvid}, such as depth, to provide additional spatial constraints during the generation process.
Due to the lack of annotated datasets for fine-grained editing training, many approaches~\cite{ceylan2023pix2video, geyer2023tokenflow,kara2024rave,khachatryan2023text2video} resorted to training-free strategies. Pix2Video~\cite{ceylan2023pix2video} used structure-guided diffusion model to edit first frame and propagate to other frames by attention in latent space. Fairy\cite{wu2024fairy} and TokenFlow\cite{geyer2023tokenflow} adopt a similar strategy of performing edits on selected frames of a video sequence. However, unlike our approach, their methods manipulate attention during the denoising steps. Moreover, their tasks are formulated as video-to-video translation, which are well-suited for stylization and appearance transfer. In contrast, our method starts from a masked video and focuses on generating ID-aligned content. The explicit use of masks in our framework provides more flexibility for editing.

\noindent\textbf{Controllable Image Generation and Inpainting.} Recent advances in controllable image synthesis~\cite{ruiz2022dreambooth,zhang2023adding,mou2023t2i} and image inpainting~\cite{li2022mat,suvorov2022resolution,dong2022incremental} have greatly enhanced the capability of generative models in visual tasks. 
ControlNet~\cite{zhang2023adding} and T2I-Adapter~\cite{mou2023t2i} introduce trainable adapters~\cite{houlsby2019parameter} that inject visual conditions into pretrained backbones for controllable generation.
Reference-guided inpainting methods~\cite{zhou2021transfill,zhao2022geofill,zhao20223dfill,cao2024leftrefill,tang2024realfill,oh2019onion} leverage additional reference images from different viewpoints to inform the completion of target image.
Among them, Leftrefill~\cite{cao2024leftrefill} explores the use of prompt tuning for improving identity consistency in reference inpainting. While effective in maintaining visual fidelity across views, the generation of Leftrefill can not controlled by language prompts and also fails to generalize to the video domain of dynamic scene.
\section{Methodology}

\begin{figure}
    \begin{center}
    \includegraphics[width=\linewidth]{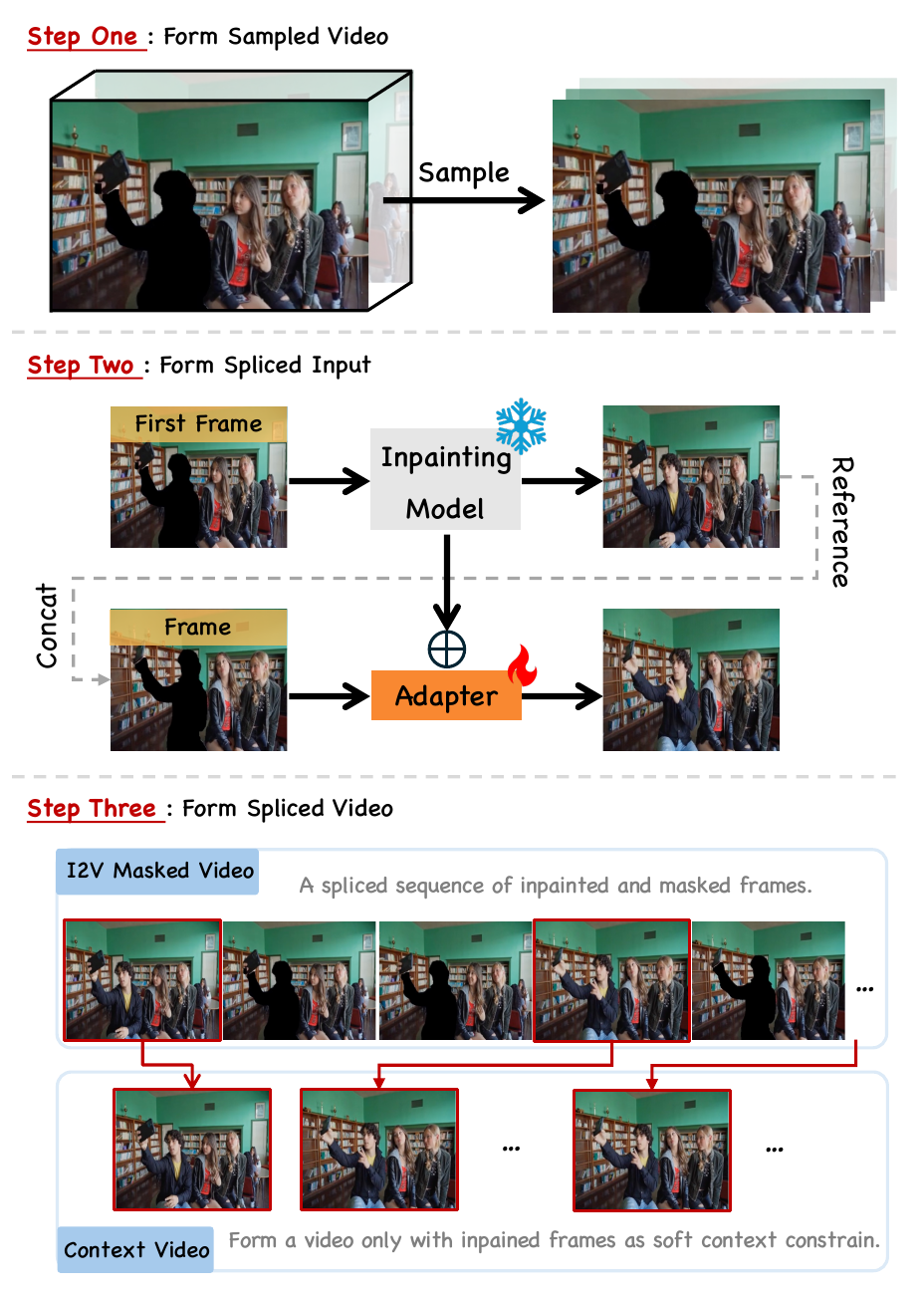}
    \end{center}
   \caption{The pipeline of the FrameProp module. It generates both the I2V-masked video and the context video through three steps: Sampled Video Formation, Reference-guided Inpainting, and Spliced Video Formation.}\label{image:FrameProp_pipe}
\end{figure}

\label{section:overall_pipe}
\begin{figure*}
\begin{center}
\includegraphics[width=\linewidth]{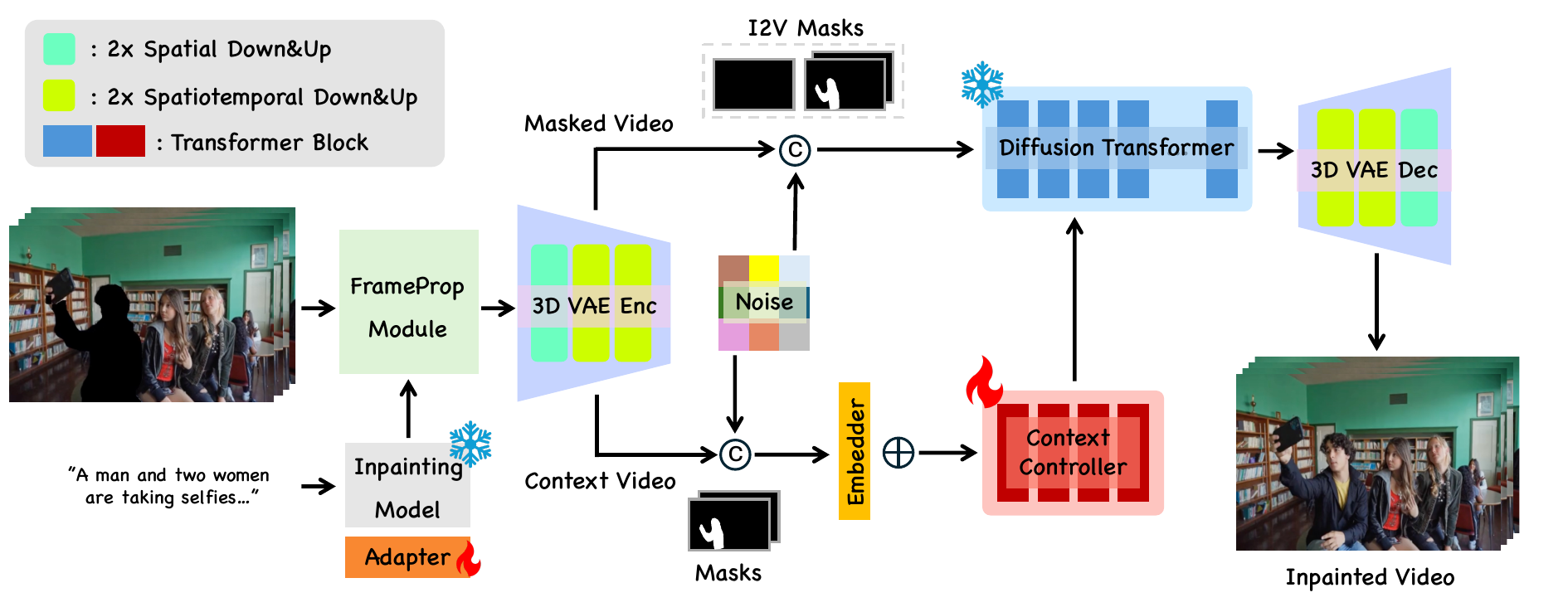}
\end{center}
   \caption{The overall pipeline of VidPivot. As shown in the figure, VidPivot consists of three main components: the FrameProp module, the context controller, and the video diffusion transformer. The FrameProp module generates ID-aligned anchor frames, while the context controller extracts the upstream-constructed context video and injects it into the video diffusion transformer to guide temporal modeling.}\label{image:pipeline}
\end{figure*}

\noindent\textbf{Problem Formulation.} Given a video sequence $ V = \left \{ {V}_{t} \in {\mathbb{R}}^{H \times W \times 3} \right \}^{T}_{t=1}$ of $T$ frames, along with corresponding binary mask for each frame $ \mathbf{M}= \left \{ {M}_{t} \in {\mathbb{R}}^{H \times W \times 1} \right \}^{T}_{t=1}$, the objective of video inpainting is to generate or remove specific elements with the masked video $ \widetilde{V} = \left \{ {V}_{t} \odot \left (1 - {M}_{t} \right ) \in {\mathbb{R}}^{H \times W \times 3} \right \}^{T}_{t=1}$ as input, guided by a textual or visual prompt $P$. The inpainted region is excepted to be spatially seamless and temporally consistent throughout the entire sequence. 

\noindent\textbf{Overview.}  Fig.~\ref{image:pipeline} provides a overview of VidPivot. Specifically, VidPivot takes a masked video $\widetilde{V}$ and a textual prompt $P$ as inputs.
The masked video is first processed by the carefully designed FrameProp Module (Sec.~\ref{section:FrameProp_module}), which produces two distinct spliced video streams: the I2V-masked video and the context video. Each stream is constructed using priors derived from an image inpainting model and incorporates identity alignment across frames to ensure instance-level temporal consistency.
Then 3D VAE encoder is employed to produce 8$\times$ spatiotemporal downsampled latent features for these two video streams.
We employ a context controller to extract contextual feature from the context video, serving as constraint of appearance preservation (Sec.~\ref{section:motion_controller}). Finally, a diffusion transformer accepts the I2V-masked video as input, guided by the contextual feature, and produce temporally coherent inpainting results (Sec.~\ref{section:motion_propagation}).
So, VidPivot's design offers key advantages for video inpainting:
\begin{itemize}
    \item Reference-guided context image inpainting can successfully preserve the identity consistency and avoid the appearance degradation problem in video inapinting task.
    \item Guided by the consistent context video, video diffusion model can focus on the problem of short-range dynamics modeling, enabling improved stability and coherence.
    \item Benefited from the decomposition of appearance preservation and motion modeling, VidPivot can generate vivid dynamics with consistent identity. 
\end{itemize}

\subsection{Multi-frame Consistent Inpainting}\label{section:FrameProp_module}

Prior work~\cite{bian2025videopainter} has demonstrated that inpainting the first frame of a video and subsequently modeling the motion within the masked regions using an I2V generation model yields more effective and higher-quality results than directly performing inpainting across the entire sequence~\cite{zi2025cococo}. Inspired by this insight, we extend the first-frame-inpainting paradigm by proposing FrameProp Module to inpaint contextual frame intervals, which diffuses the content of initial reference frame into subsequent sampled frames through a splicing mechanism. Specifically, the FrameProp Module consists of three stages: sampled video formation, reference-guided inpainting and spliced video formation.

\noindent\textbf{Sampled Video Formation.} While a full video sequence contains rich contextual information, it also contains unnecessary redundancy for video inpainting task. To address this, we apply a temporal sampling strategy with a stride of $K$, selecting $\lceil T / K\rceil$ key frames from the original sequence. These key frames capture salient inter-frame variations in viewpoint and scene dynamics, thereby providing informative guidance for subsequent video inpainting while remaining compactness. 

\noindent\textbf{Reference-guided Inpainting.} After temporal sampling, we perform inpainting on both the first frame and the sampled frames. Due to viewpoint variations across frames, inpainting all frames independently leads to discrepancies in structure and appearance within frames. As illustrated in \textbf{Step Two} of Fig.~\ref{image:FrameProp_pipe}, we introduce a novel splicing mechanism to enforce identity alignment across frames. Specifically, we pretrain a low-rank ID adapter $\phi$ on image inpainting model $\Psi$ for ID alignment:
\begin{equation}
    h = Wx+B_{\phi}A_{\phi}x, \quad B_{\phi}\in\mathbb{R}^{d\times r},A_{\phi}\in\mathbb{R}^{r\times d},
\end{equation}
where $x$ and $h$ indicate input and output features with channel $d$ for the linear layers in attention blocks; $W$ denotes the frozen weights of original inpainting model, while $B_{\phi},A_{\phi}$ are trainable low-rank matrices with much fewer parameters compared to $W$, \textit{i.e.}, $r \ll d$.
Once the first frame is inpainted, it is concatenated along the width dimension inspired by~\cite{cao2024leftrefill} with each sampled frame and fed into the image inpainting model with adapter. This spliced pixel-level generation mechanism is highly effective for appearance preservation, as shown in Fig.~\ref{image:teaser}, where the reference context flows naturally into the target masked region. The ID adapter is trained on a multi-view dataset and can be broadly applicable to various downstream tasks. The training process can be formulated as follows:
\begin{equation}
\mathcal{L}_{ID}(\phi) = \mathbb{E}\left[ \| (\varepsilon - x_0) - \Psi_\phi(\left[x_t;\widetilde{x}_0; M \right], P, t) \|^2 \right]
\end{equation}
where $\varepsilon$ is noise, $x_0$ is the groundtruth, $\widetilde{x}_0$ is masked frame latent, $x_t$ is noised frame latent, and $M$ is image mask.
Furthermore, benefited from the strong associative capabilities of state-of-the-art image inpainting model, our method effectively injects novel view synthesis ability into the VidPivot pipeline. As illustrated in the red car example in Fig.~\ref{image:teaser}, VidPivot is able to plausibly infer and synthesize unseen regions from new viewpoints, significantly enhancing spatial consistency and realism. Notably, in scenarios involving long-time motion, we inpaint the first frame and then partition the video into several temporal chunks. For each identity-aligned chunk, the last inpainted frame is used as first frame for guiding the inpainting of the subsequent chunk to keep inter-chunk consistency.

\noindent\textbf{Spliced Video Formation.} The contextual frame intervals are utilized as anchors to construct the new I2V masked video ${V}_{m}$ and context video ${V}_{c}$, as illustrated in \textbf{Step Three} of Fig.~\ref{image:FrameProp_pipe}. For the I2V-masked video, we directly insert the inpainted frames back into their original temporal positions, while leaving the remaining masked frames unchanged. Besides, we construct I2V video mask $M^{i2v}$ with the original input mask, expect the mask of first frame which is set to all zero. The resulting sequence and mask serve as the input to the I2V branch. It should be notice that a mask value of 1 indicates a region that can be modified, while a value of 0 denotes fixed content. Therefore, the inpainted frame intervals serve as a soft constraint to guide the generation process, which remains adaptable by the model to ensure superior temporal continuity. 
In parallel, we construct a context video as input for context controller, which in turn guides the generation process of the backbone network. This context video is formed by directly assembling the inpainted frame intervals into a new, condensed sequence, paired with a sequence of corresponding mask $M^{0}$ . 

\noindent\subsection{Context Guidance via Controller Injection}
\label{section:motion_controller}
As the I2V backbone natively lacks the capability for inpainting, we introduce a context controller to guide the generation of diffusion model with context video, thereby empowering the model to accomplish these tasks. The controller is initialized with pretrained diffusion transformer blocks. Inspired by~\cite{bahmani2024ac3d,liang2024wonderland}, VidPivot context controller injects control information only into the first 10 layers of the I2V diffusion transformer, which captures low-level camera and motion information. For the context video $V_{c}$ and I2V masked video $V_m$ generated in Sec.~\ref{section:FrameProp_module}, we first employ a 3D VAE encoder to get their latent features $f_c$ and $f_m$, respectively. The latent feature is then concatenated along the channel dimension with masked video latent and corresponding video masks. Following ControlNet~\cite{zhang2023adding}, the feature of context video is then passed through a zero-initialized learnable embedding layer $\mathcal{F}$. The context controller $\mathcal{H}$ accepts this zero-initialed embedded feature and embedding of I2V video as input and get the control feature $f$ as follows: 
\begin{equation}
f = \mathcal{H}\left(\mathcal{F} \left(\left [ f_{c}; M^{0} \right ]\right) + \mathcal{F}_{m}\left( \left [ f_{m}; M^{i2v} \right ]\right)\right),
\label{eq:injection_control}
\end{equation}
where $\left [ \cdot \right ]$ denotes concatenation operation, $\mathcal{F}_{m}$ is the pretrained embedding layer of I2V branch.
Our formulation ensures that the context latent captures motion variations across frame intervals while remaining stable throughout the forward diffusion process.

\subsection{Masked Area Content Propagation}
\label{section:motion_propagation}
For the I2V masked video generated in Sec.~\ref{section:FrameProp_module}, we similarly concatenate it with randomly initialized noise and its corresponding mask along the channel dimension and send it to diffusion transformer $\mathcal{D}$. 
After obtaining the context feature $f_c$ from Eq.~\ref{eq:injection_control}, we add it into the output feature of corresponding layers of DiT during the forward pass. We employ flow matching loss~\cite{lipman2022flow} for the joint optimization of context controller while freeze the diffusion transformer:
\begin{align}
\mathcal{L}_{video}(\mathcal{H}, \mathcal{F}) = \left[ \left\| (\varepsilon - \hat{f}^0) - \mathcal{D}\left(\left[\hat{f}^t;f_m;M^{i2v} \right], f\right) \right\|^2 \right],
\end{align}
where $\hat{f}^0$ is the latent feature of groundtruth video $V$, $\hat{f}^t$ represents the noised video latent of $\hat{f}^0$.

\section{Experiments}
\label{section:experiments}
\textbf{Setups}.
VidPivot is built upon the Wan~\cite{wang2025wan} I2V-1.3B/14B. The training process consists of two stages:
In the first stage, we train an ID alignment adapter based on FLUX.Fill~\cite{fluxfill2024} with a LoRA rank of 128 at a resolution of 512$\times$ 512. We use the AdamW optimizer with a learning rate of 1e-4, a batch size of 16, and train for 80,000 steps. The sampling step is set to 50 to enhance performance across different inpainting tasks.
In the second stage, we construct the FrameProp module using the pretrained adapter and proceed to train the transformer and the context controller jointly. This stage is conducted at a resolution of 480P, with a batch size of 1, using AdamW with a learning rate of 1e-5. We utilize 8 H100 GPUs to run all experiments.

\noindent\textbf{Dataset.} To enable effective ID alignment in the FrameProp module, we utilize a multi-view dataset. Specifically, we use image pairs from MegaDepth\cite{li2018megadepth}, which contains diverse multi-view scenes of landmarks. For training, masks are constructed from a combination of randomly generated masks and semantic masks from the COCO dataset with a masking ratio ranging from 40\% to 70\%. In total, we curate a dataset of 820K masked image pairs for training the adapter. For training the embedding layer and context controller, we select the VPData dataset introduced by VideoPainter~\cite{bian2025videopainter}, which consists of 450K publicly available internet videos covering a wide range of indoor and outdoor scenes. For evaluation, we conduct experiments on both the widely used DAVIS dataset~\cite{perazzi2016benchmark} and the VPBench benchmark (VPBench-S and VPBench-L) proposed by VideoPainter.

\noindent\textbf{Metrics.} We evaluate the perseverance of the unmasked regions in the generated video using metrics including PSNR, SSIM, LPIPS, and MSE. For a clearer comparison of the results, we scale both LPIPS and MSE metrics 100$\times$ for convenient comparison. To assess the semantic alignment between the generated content within the masked regions and the textual prompt, we employ the CLIP similarity~\cite{radford2021learning} score as a quantitative measure. We also evaluate FVD metric for overall video temporal quality.

\subsection{Results of Video Inpainting}
\begin{figure*}
\includegraphics[width=\linewidth]{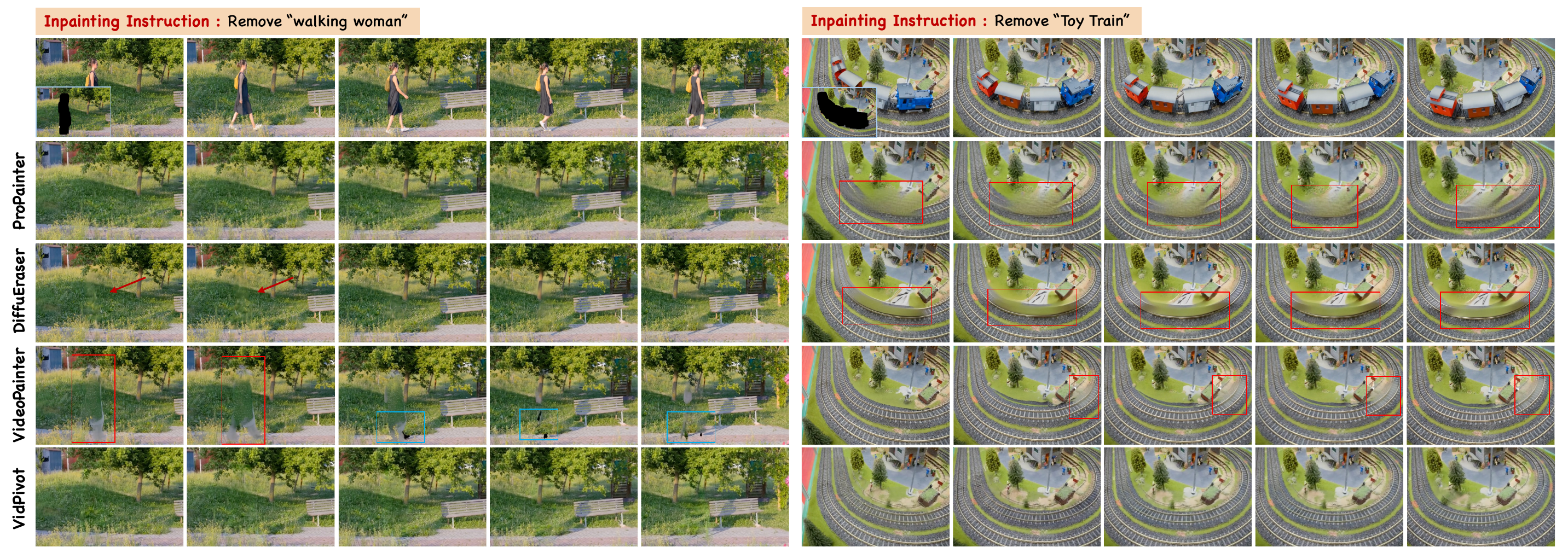}
   \caption{The inpainting comparison between VidPivot, ProPainter, DiffuEraser and VideoPainter.}\label{image:inpaintingcompare}
\end{figure*}

\textbf{Quantitative results.}
\begin{table*}[t]
  \centering
  \footnotesize 
  \setlength{\tabcolsep}{3pt}
  \begin{tabular}{cc|cccccc|c|cccccc}
    \toprule
    & \multicolumn{7}{c|}{\textbf{Video Inpainting}} & \multicolumn{7}{c}{\textbf{Video Editing}} \\
    \midrule
    & \textbf{Models} & PSNR$\uparrow$ & SSIM$\uparrow$ & LPIPS$\downarrow$ & MSE$\downarrow$  & CLIP Sim$\uparrow$ & FVD$\downarrow$ & \textbf{Models} & PSNR$\uparrow$ & SSIM$\uparrow$ & LPIPS$\downarrow$ & MSE$\downarrow$  & CLIP Sim$\uparrow$ & FVD$\downarrow$\\
    \midrule
    \multirow{6}{*}{\rotatebox{90}{VPBench-S}} 
    & ProPainter   & 20.97 & \cellcolor{rank2}0.87 & 9.89 & 1.24 & 17.18 & 0.44 & UniEdit   & 9.96 & 0.36 & 56.68 & 11.08 & 14.23 & 1.36\\
    & COCOCO       & 19.27 & 0.67 & 14.80 & 1.62 & 20.03 & 0.69 & DitCtrl  & 9.30 & 0.33 & 57.42 & 12.73 & 15.59 & 0.57\\
    & Cog-Inp      & 22.15 & 0.82 & 9.56 & 0.88 & \cellcolor{rank2}21.27 & \cellcolor{rank3}0.18 & ReVideo   & 15.52 & 0.49 & 27.68 & 3.49 & \cellcolor{rank3}20.01 & 0.42\\
    & VACE & \cellcolor{rank2} 25.77 & 0.81 & \cellcolor{rank3}7.94 & \cellcolor{rank2}0.47 & 18.86 & 0.19 & VACE & \cellcolor{rank2}26.14 & 0.80 & \cellcolor{rank2}7.63 & \cellcolor{rank2}0.49 & 17.02 & \cellcolor{rank3}0.20\\
    & VideoPainter & \cellcolor{rank3}23.32 & \cellcolor{rank1}0.89 & \cellcolor{rank2}6.85 & \cellcolor{rank3}0.82 & \cellcolor{rank1}21.49 & \cellcolor{rank1}0.15 & VideoPainter & \cellcolor{rank3}22.63 & \cellcolor{rank1}0.91 & \cellcolor{rank3}7.65 & \cellcolor{rank3}1.02 & \cellcolor{rank2}20.20 & \cellcolor{rank2}0.18 \\
    & VidPivot    & \cellcolor{rank1}28.22    & \cellcolor{rank3}0.85    & \cellcolor{rank1}5.95    & \cellcolor{rank1}0.24    & \cellcolor{rank3}21.24     & \cellcolor{rank1}0.15 & VidPivot    & \cellcolor{rank1}28.73     & \cellcolor{rank2}0.86    & \cellcolor{rank1}6.86    & \cellcolor{rank1}0.20    & \cellcolor{rank1}20.64 & \cellcolor{rank1}0.16 \\
    \midrule
    \multirow{6}{*}{\rotatebox{90}{VPBench-L}} 
    & ProPainter   & 20.11 & \cellcolor{rank2}0.84 & 11.18 & 1.17 & 17.68 & 0.48 & UniEdit   & 10.37 & 0.30 & 54.61 & 10.25 & 15.42 & 1.00\\
    & COCOCO       & 19.51 & 0.66 & 16.17 & 1.29 & 20.42 & 0.62 & DitCtrl   & 9.76 & 0.28 & 62.49 & 11.50 & 16.52 & 0.56 \\
    & Cog-Inp      & 19.78 & 0.73 & 12.53 & 1.33 & \cellcolor{rank3}21.22 & \cellcolor{rank2}0.21 & ReVideo   & 15.50 & 0.46 & 28.57 & 3.92 & \cellcolor{rank1}20.50 & 0.35\\
    & VACE & \cellcolor{rank2}24.57 & \cellcolor{rank3}0.75 & \cellcolor{rank3}10.35 & \cellcolor{rank3}1.00 & 20.45 & 0.29 & VACE & \cellcolor{rank3}21.70 & \cellcolor{rank3}0.70 & \cellcolor{rank3}12.36 & \cellcolor{rank3}1.12 & 17.03 & \cellcolor{rank3}0.27\\
    & VideoPainter & \cellcolor{rank3}22.19 & \cellcolor{rank1}0.85 & \cellcolor{rank2}9.14 & \cellcolor{rank2}0.71 & \cellcolor{rank2}21.54 & \cellcolor{rank1}0.17 & VideoPainter & \cellcolor{rank2}22.60 & \cellcolor{rank1}0.90 & \cellcolor{rank1}7.53 & \cellcolor{rank2}0.86 & \cellcolor{rank3}19.38 & \cellcolor{rank1}0.11 \\
    & VidPivot    & \cellcolor{rank1}28.11    & \cellcolor{rank2}0.84    & \cellcolor{rank1}5.31    & \cellcolor{rank1}0.26  & \cellcolor{rank1}22.32  & \cellcolor{rank3}0.24  & VidPivot    & \cellcolor{rank1}26.17     & \cellcolor{rank2}0.82    & \cellcolor{rank2}8.11    & \cellcolor{rank1}0.28    & \cellcolor{rank2}19.55 & \cellcolor{rank2}0.21\\
    \bottomrule
  \end{tabular}
  \caption{Inpainting and editing evaluation on VPBench. Deeper color indicate better performance. Note we scale LPIPS and MSE metrics 100$\times$ for convenient comparison.}
  \label{tab:vpbench}
\end{table*}
As shown in Tab.~\ref{tab:vpbench} and Tab.~\ref{tab:davis_inpainting}, we conduct quantitative comparisons with state-of-the-art video inpainting methods. Here, we select official VACE-LTX as our baseline for comparison, as we empirically found it to show stronger capability at removal task than Wan version. Both VACE~\cite{yang2024cogvideox} and VideoPainter demonstrate strong generative capabilities within the masked regions, as reflected by their high CLIP similarity scores. Our proposed VidPivot also achieves competitive performance in terms of content generation, particularly on the VPBench-L benchmark. This can be attributed to the incorporation of anchor frames and context video, which effectively constrain the generation process. In addition, VidPivot attains competitive results on background preservation metrics, demonstrating its ability to maintain the integrity of unmasked content while performing high-quality foreground inpainting.

\noindent\textbf{Qualitative results.}
We visualize the qualitative results of state-of-the-art methods—ProPainter~\cite{zhou2023propainter}, DiffuEraser~\cite{li2025diffueraser}, VideoPainter~\cite{bian2025videopainter}, and our proposed VidPivot—on representative video inpainting tasks, as shown in Fig.~\ref{image:inpaintingcompare}. Benefiting from the incorporation of both inpainting priors and context priors, VidPivot exhibits stable object removal and successfully reconstructs occluded regions such as grass and rail tracks within the masked areas. ProPainter, which is based on a transformer architecture, effectively removes the target region but struggles to generate semantically consistent content in occluded areas. DiffuEraser, initialized Gaussian noise with ProPainter’s output, introduces noticeable blurriness in the inpainted regions due to imperfect mask-aware generation. VideoPainter, on the other hand, suffers from severe artifacts in removal scenarios: residual motion are still visible inside the masked area, and only partial alignment with the background is achieved, as highlighted in the red box in Fig.~\ref{image:inpaintingcompare}. Additionally, VideoPainter exhibits unfilled regions (see blue boxes), likely due to insufficient learning of its context branch. In contrast, VidPivot maintains both spatial coherence and temporal consistency, demonstrating more stable content propagation across frames.

\subsection{Results of Video Editing}
\begin{figure*}
\centering
\includegraphics[width=\linewidth]{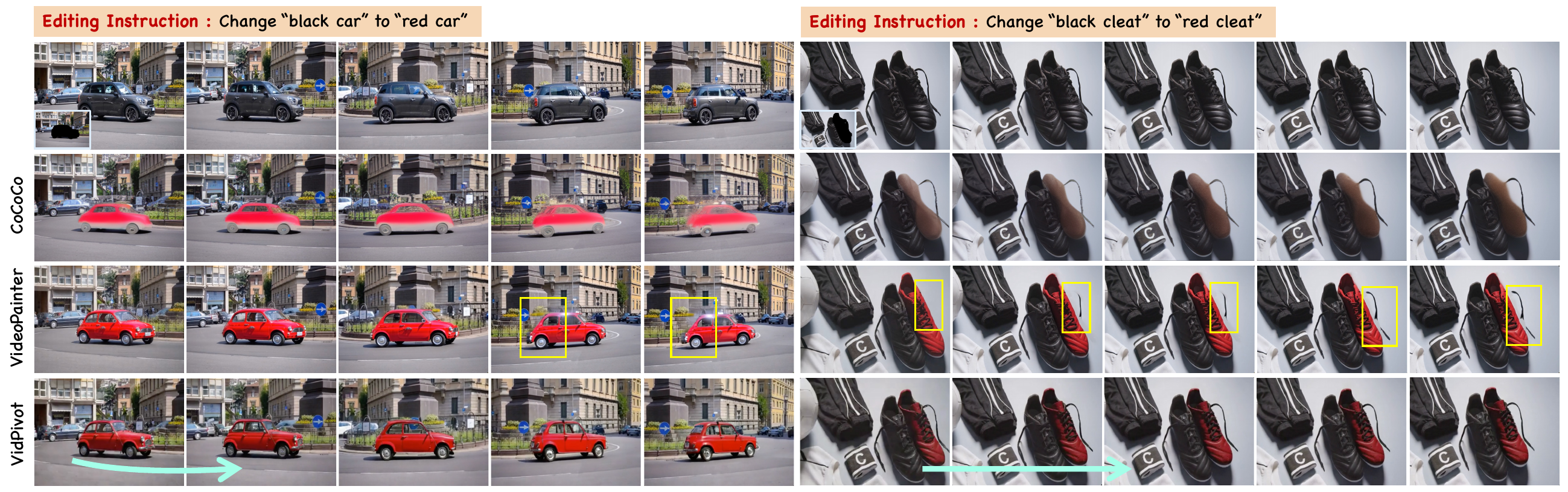}
\caption{The frame-by-frame editing comparison between VidPivot, ProPainter and COCOCO.}\label{image:editingcompare}
\end{figure*}

\noindent\textbf{Quantitative results.}
Following VideoPainter, we conduct quantitative comparisons against recent video editing models including UniEdit~\cite{bai2024uniedit}, DitCtrl~\cite{cai2024ditctrl}, and ReVideo~\cite{mou2024revideo}, as summarized in Tab.~\ref{tab:vpbench}. VidPivot consistently outperforms competitors on background preservation metrics across both short videos (VPBench-S) and long videos (VPBench-L), with particularly strong performance in PSNR. VideoPainter benefits from training on higher-resolution videos, which gives it great SSIM metrics. Notably, VidPivot still achieves top-tier CLIP similarity scores, highlighting the effectiveness of injected contextual frame intervals in enhancing semantic alignment with the prompt. In summary, VidPivot achieves state-of-the-art quantitative performance in both semantic alignment of foreground generation and background consistency preservation.

\begin{table*}[t]
  \centering
  \scriptsize
  \setlength{\tabcolsep}{3pt}
  \begin{minipage}[t]{0.5\linewidth}
    \centering
    \begin{tabular}{cc|cccccc}
      \toprule
      & \textbf{Models} & PSNR$\uparrow$ & SSIM$\uparrow$ & LPIPS$\downarrow$ & MSE$\downarrow$  & CLIP Sim$\uparrow$ & FVD$\downarrow$\\
      \midrule
      \multirow{7}{*}{\rotatebox{90}{DAVIS Removal}} 
      & COCOCO       & 21.34 & 0.66 & 10.51 & \cellcolor{rank3}0.92 & 17.50 & 0.33\\
      & Cog-Inp      & 23.92 & \cellcolor{rank2}0.79 & 10.78 & \cellcolor{rank2}0.47 & \cellcolor{rank3}17.53 & \cellcolor{rank1}0.17\\
      \cmidrule(lr){2-8}
      & ProPainter   & \cellcolor{rank2}26.05 & 0.73 & \cellcolor{rank3}9.90 & 1.21 & 15.02 & 0.29 \\
      & VACE     & 22.80 & 0.70 & 13.43 & 1.34 & \cellcolor{rank2}19.25 & \cellcolor{rank3}0.28 \\
      & DiffuEraser & \cellcolor{rank3}25.21 & \cellcolor{rank3}0.73 & \cellcolor{rank2}9.72 & 1.22 & 15.25 & 0.28 \\
      & VidPivot   & \cellcolor{rank1}27.34 & \cellcolor{rank1}0.82 & \cellcolor{rank1}9.15 & \cellcolor{rank1}0.25 & \cellcolor{rank1}20.54 & \cellcolor{rank2}0.19\\
      \bottomrule
    \end{tabular}
    \caption{Inpainting evaluation on DAVIS removal task.}
    \label{tab:davis_inpainting}
  \end{minipage}%
  \hfill
  \begin{minipage}[t]{0.5\linewidth}
    \centering
    \begin{tabular}{c|ccccc}
      \toprule
      \textbf{Models} & PSNR$\uparrow$ & SSIM$\uparrow$ & LPIPS$\downarrow$ & MSE$\downarrow$  & CLIP Sim$\uparrow$ \\
      \midrule
      Stride=5   & 26.57 & 0.81 & 8.57 & 0.30 & 20.48 \\
      Stride=20  & 26.52 & 0.81 & 8.99 & 0.30 & 20.49 \\
      Layers=2   & 22.05 & 0.75 & 11.01 & 0.81 & 20.34 \\
      Layers=5   & 26.34 & 0.81 & 8.66 & 0.32 & 20.54  \\
      w/o ID Align & 25.23 & 0.78 & 10.95 & 0.59 & 20.87 \\
      VidPivot & 27.34 & 0.82 & 9.15 & 0.25 & 20.54 \\
      \bottomrule
    \end{tabular}
    \caption{Ablation study evaluation on DAVIS.}
    \label{tab:ablation_study}
  \end{minipage}
  
  \vspace{-10pt}
\end{table*}
\noindent\textbf{Qualitative results.} 
We compare VidPivot with representative video editing methods that accept masked video inputs, namely COCOCO and VideoPainter, on both VPBench and DAVIS, as visualized in Fig.~\ref{image:editingcompare}. The selected examples represent two common scenarios: one involving object motion, and the other involving camera motion. Similar to VideoPainter, VidPivot leverages the inpainted first frame as a strong anchor, achieving high-quality edits at the beginning of the video. In contrast, although COCOCO produces semantically correct edits, its results often suffer from inferior visual quality and temporal inconsistency across the entire video. In the “red car” case from Fig.~\ref{image:editingcompare}, VideoPainter exhibits content degradation toward the end of the video. In contrast, VidPivot maintains high fidelity, thanks to its multi-frame consistent inpainting priors introduced by the FrameProp module, which enable more robust generation even under large object motions and novel view transitions. In the “red cleat” case, VidPivot demonstrates stable inpainting performance, while VideoPainter exhibits temporal flickering, particularly around the “shoelace” region. We also summarize the Fréchet Video Distance (FVD) scores for overall video quality and motion smoothness in Tab.~\ref{tab:vpbench} and Tab.~\ref{tab:davis_inpainting}, which demonstrate that VidPivot, VideoPainter and VACE achieves competitive generative performance within the masked region.

\subsection{Ablation Analysis}
In our proposed VidPivot framework, the sample stride serves as a key hyperparameter that determines how many contextual frame intervals are pre-inpainted and injected into the video generation process. Another crucial parameter is the number of controller layers, which directly influences the model’s capacity to capture camera and motion information. Additionally, we conduct ablation on the FrameProp Module to evaluate the effectiveness of the ID alignment strategy introduced for enhancing frame consistency. All results are summarized in Tab.~\ref{tab:ablation_study}.

\begin{figure}
    \begin{center}
    \includegraphics[width=\linewidth]{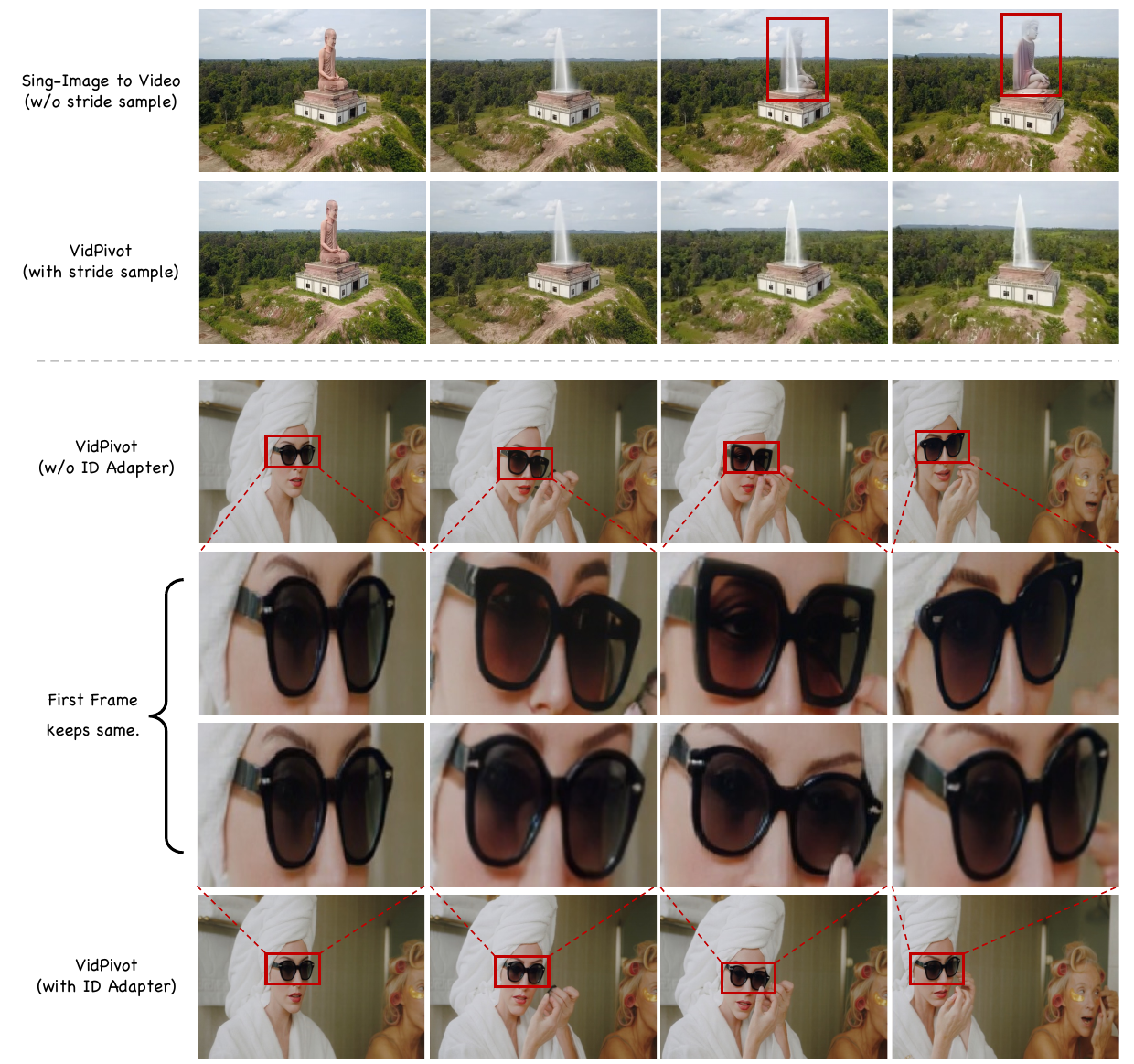}
    \end{center}
   \caption{Qualitative ablation results of VidPivot framework.}\label{image:ablation_vis}
   \vspace{-10pt}
\end{figure}

\noindent\textbf{Sample Stride.} As shown in row 1, 2, 6 and Fig.~\ref{image:ablation_vis}, single-image or few-image propagation leads to acceptable background preservation but causes a significant content drop in some cases. An over-reliance on image priors (small stride) can compromise visual fidelity. This demonstrates that selecting a proper stride can achieve not only high efficiency but also superior performance. VidPivot strikes such balance.

\noindent\textbf{Controller Layers.} As shown in the 3rd, 4th, and 6th rows of Tab.~\ref{tab:ablation_study}, reducing the number of context controller layers leads to a sharp drop in performance, demonstrating the effectiveness of our controller design. This can be attributed to fewer injection layers limit the I2V model’s ability to learn context and motion information.

\noindent\textbf{FrameProp Modulde.} When the ID alignment adapter is removed and unpainted frames are directly fed input into the image inpainting model, we observe an increase in CLIP similarity in Tab.~\ref{tab:ablation_study} row 5. However, this comes at the cost of significantly reduced temporal consistency in the output video and detailed qualitative results of ID alignment can be found in Fig.~\ref{image:ablation_vis}. This provides strong evidence for the effectiveness of our FrameProp module, which ensures consistent foreground inpainting.

\subsection{Discussion about Challenging Scenarios}
\begin{figure*}
\centering
\includegraphics[width=\linewidth]{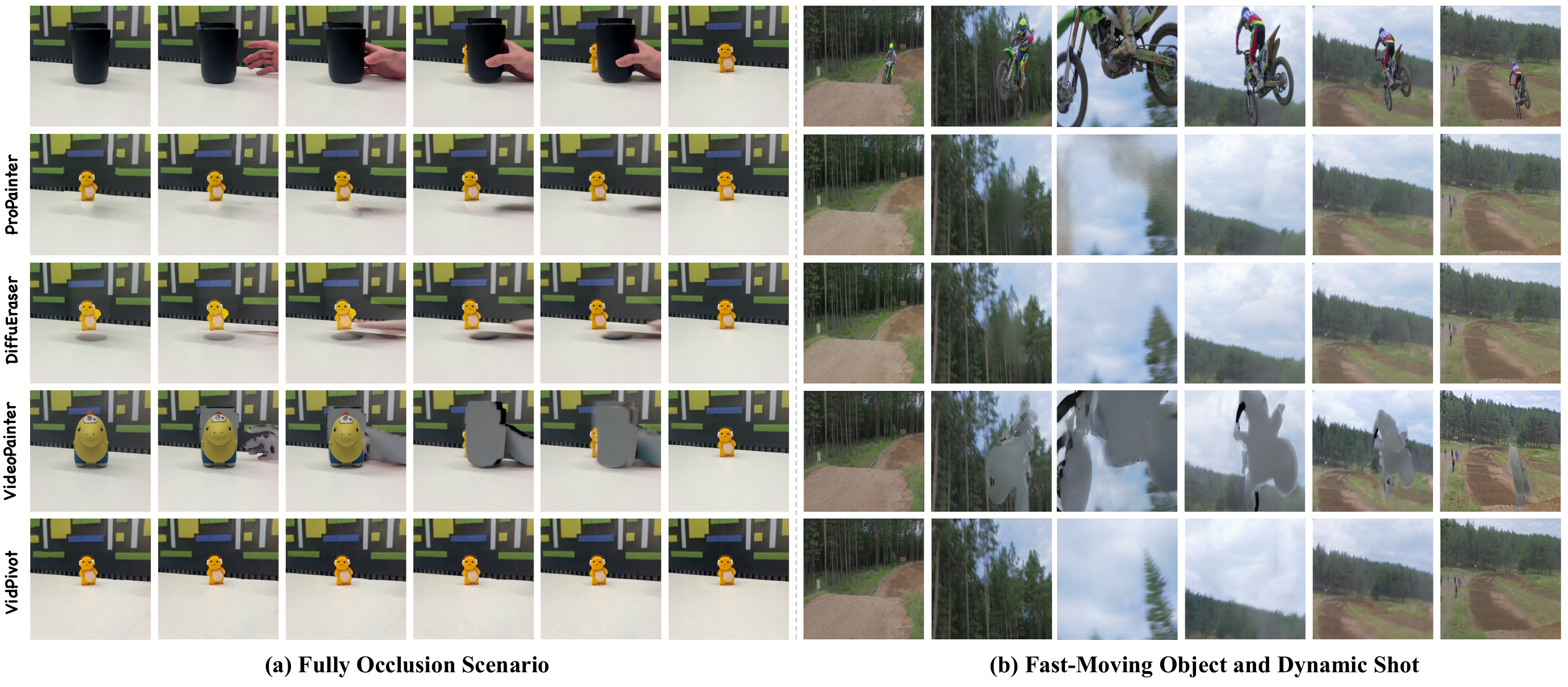}
   \caption{More challenging and practical qualitative visualizations of (a) fully occlusion scenario, (b) fast-moving object with dynamic shot.}\label{image:occ_and_fast}
\end{figure*}

\textbf{Fully Occlusion Scenario.}
Image-to-Video (I2V) models, such as VideoPainter, are capable of addressing a wide range of video editing tasks by utilizing an edited image and a corresponding mask. However, these models often falter in object removal tasks, especially when encountering scenarios with full occlusion. This frequently results in inference failure due to a lack of frame-to-frame consistency. 
DiffuEraser uses the output from ProPainter to initialize the diffusion noise. 
Our VidPivot also can easily integrate Propainter pixel warping results to serve subsequent reference inpainting, ensuring effective propagation of identity information and thereby maintaining inter-frame consistency. We present a visual comparison in Fig.~\ref{image:occ_and_fast} (a) that demonstrates the performance of different optical flow-based methods in a fully occluded scenario.

\noindent\textbf{Fast-Moving Object and Dynamic Shot Scenario.}
Another challenging scenario in practical applications is the presence of fast-moving objects and dynamic shots, which demands superior temporal stability and scene consistency from the model. We use the ``motocross-jump" sequence from the DAVIS dataset as an example to evaluate performance under these demanding conditions. This scene features a motorcycle executing a high-speed jump, coupled with significant camera motion. As illustrated in Figure~\ref{image:occ_and_fast} (b), we also provide a comparative analysis of the inference results from ProPainter, DiffuEraser, VideoPainter and our VidPivot. Both ProPainter and DiffuEraser can achieve great or acceptable object removal results, while VideoPainter often exhibits inferior inpainting performance when confronted with such dynamic scenarios. This limitation in VideoPainter can be attributed to the lack of contextual information injection faces with complex physical and motion modeling. Our VidPivot achieves competitive object removal results without inference time optical flow warping, even in this highly dynamic scenario.
\section{Conclusion}\label{section:conlcusion}
This paper proposes VidPivot, a novel video inpainting framework that decouples spatiotemporal modeling into two specialized sub-tasks: multi-frame inpainting and masked area propagation. It leverages a FrameProp module and a lightweight context controller to explicitly harness frame and motion cues, enhancing the generative process. The FrameProp module ensures content alignment across frames, preserving foreground fidelity even amidst significant motion. Unlike conventional methods that rely on single-frame input and implicit motion learning, VidPivot enables precise content control. Collectively, these components elevate overall inpainting quality and mitigate distortion across diverse video scenes.
{
    \small
    \bibliographystyle{ieeenat_fullname}
    \bibliography{main}
}

\clearpage
\setcounter{page}{1}
\maketitlesupplementary

\appendix


\section{More qualitative results of VidPivot}
\label{section:supp_qual}
\begin{figure*}
\centering
\includegraphics[width=0.76\linewidth]{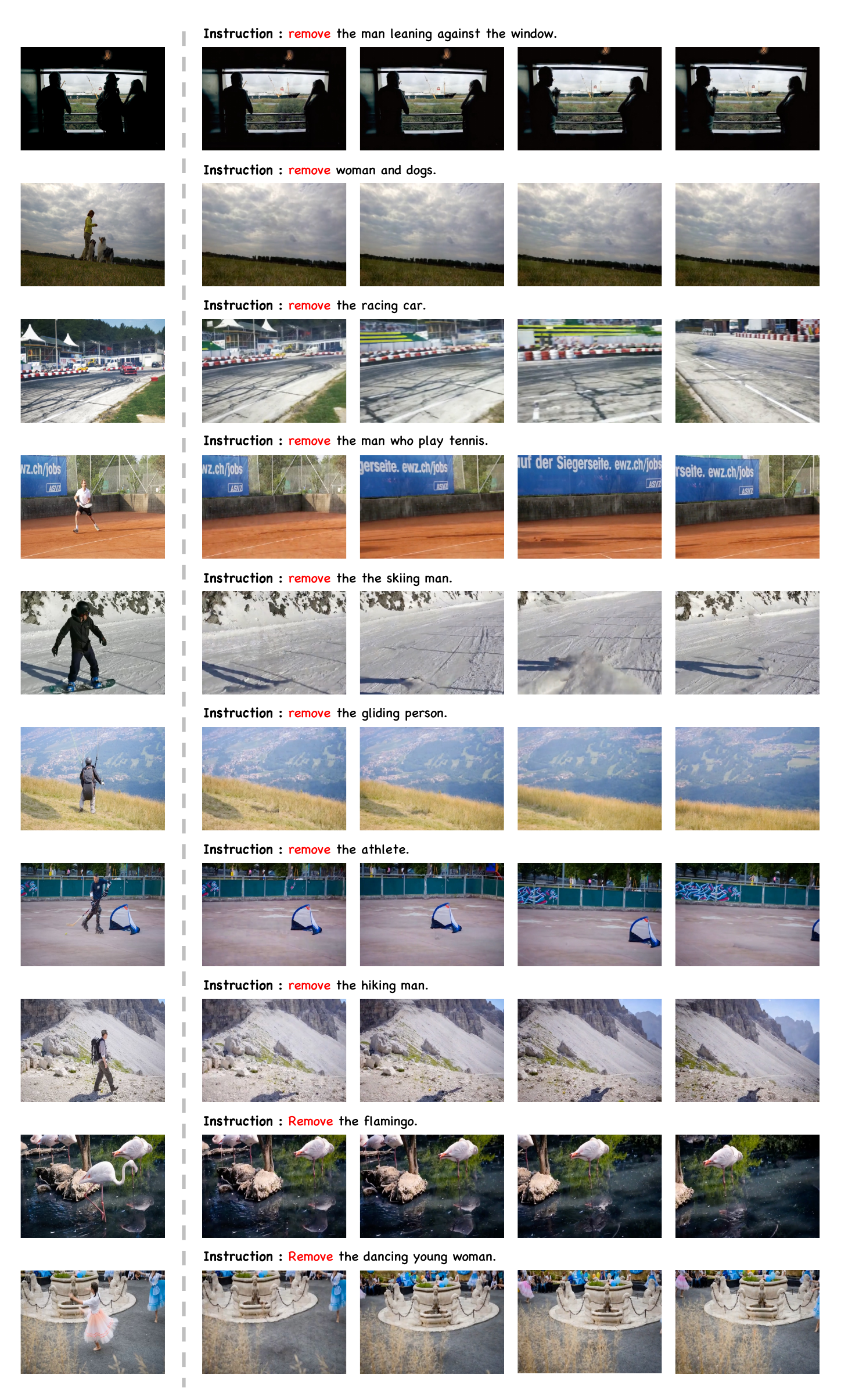}
\caption{The frame-by-frame qualitative results of VidPivot.}\label{image:supp_inpaint}
\end{figure*}

\begin{figure*}
\centering
\includegraphics[width=0.76\linewidth]{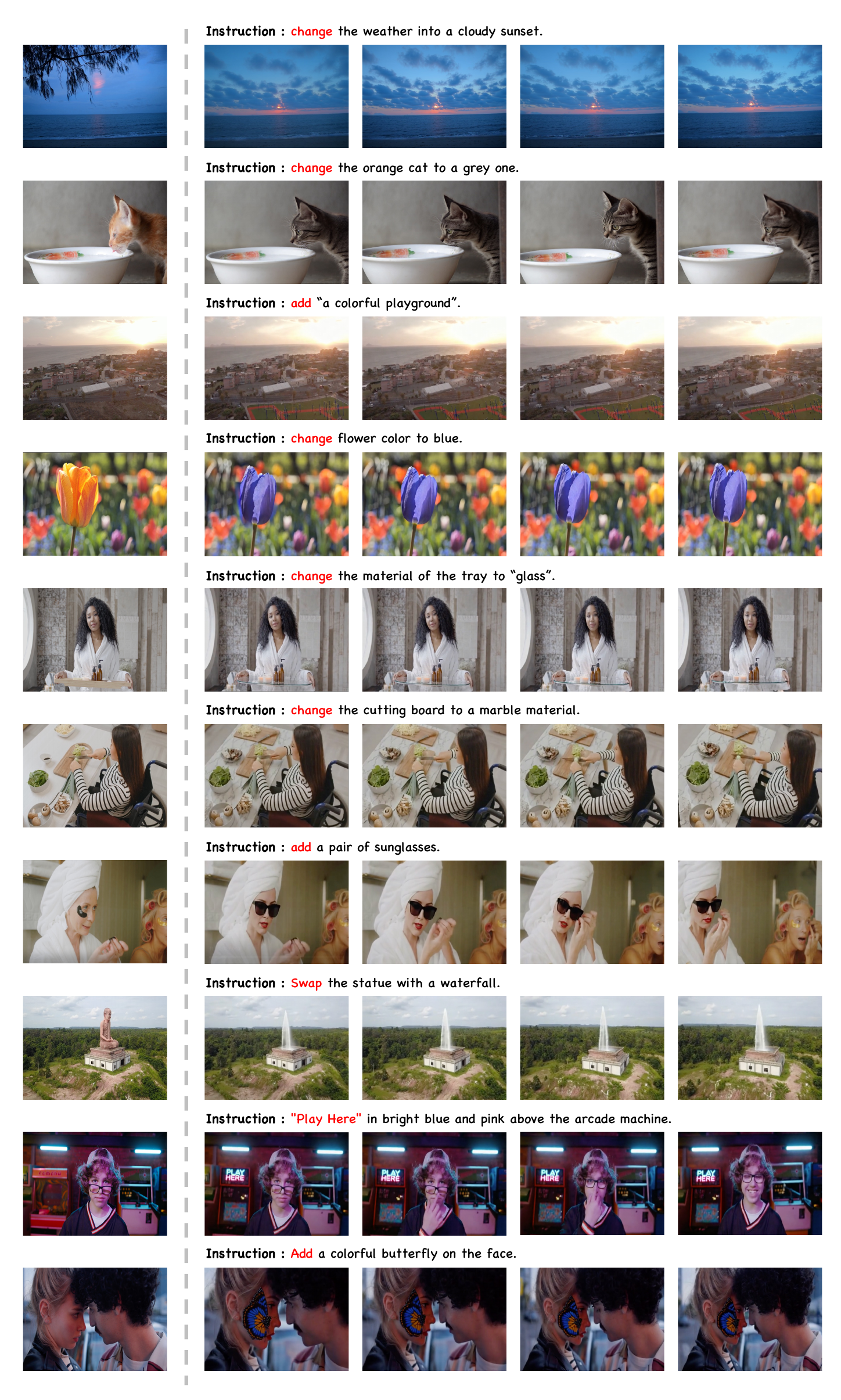}
\caption{The frame-by-frame qualitative results of VidPivot.}\label{image:supp_edit}
\end{figure*}

\begin{figure*}
\centering
\includegraphics[width=0.74\linewidth]{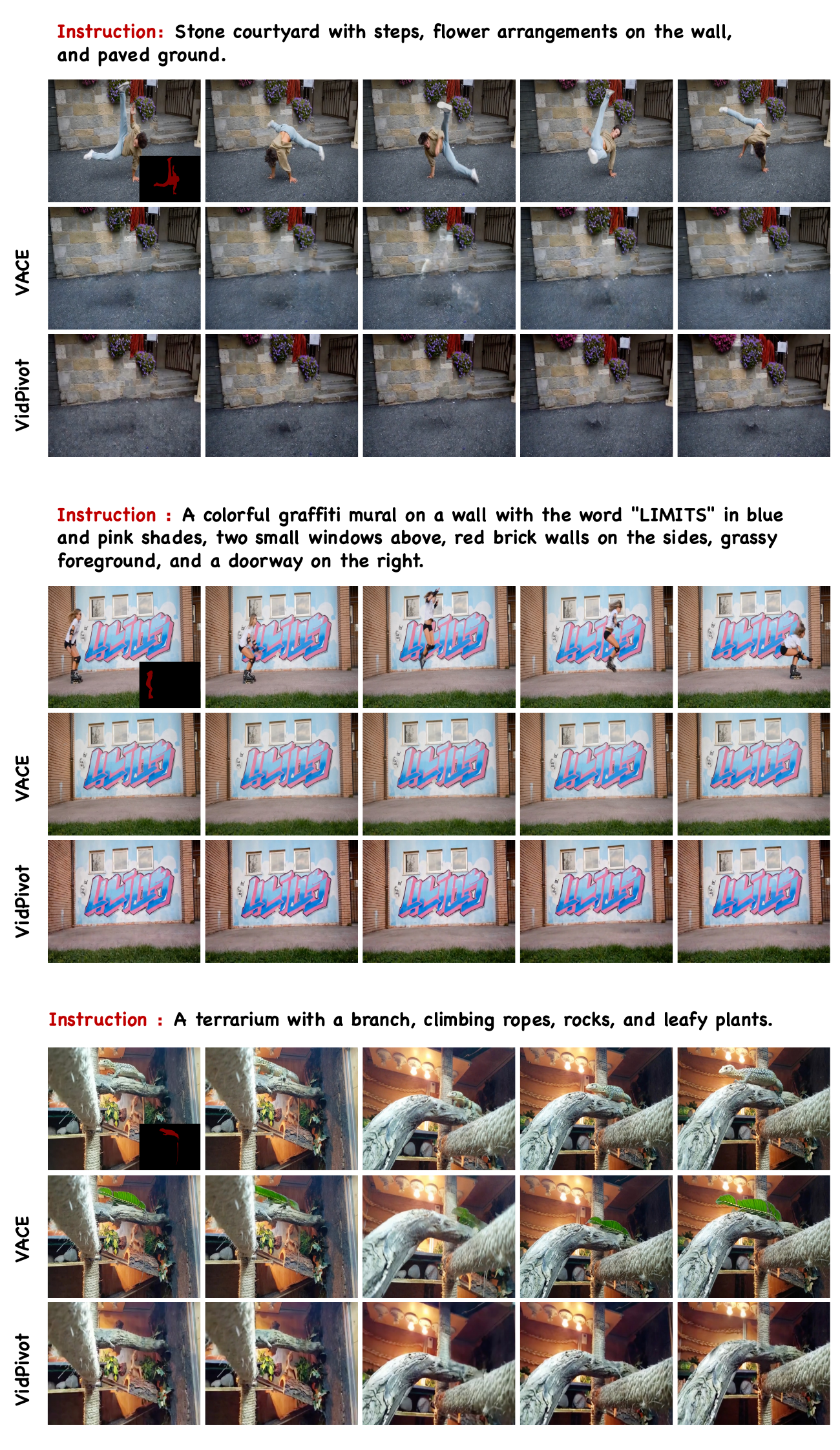}
\caption{The frame-by-frame comparison between VidPivot and VACE.}\label{image:supp_comp_vace}
\end{figure*}

Fig.~\ref{image:supp_inpaint} and Fig.~\ref{image:supp_edit} presents additional qualitative results of VidPivot. The first column shows the first frame of the original video, and the remaining columns depict generated frames. 
We select multiple samples from the DAVIS and VPBench dataset to perform inpainting and object removal.
As shown in Fig.~\ref{image:supp_inpaint}, VidPivot achieves effective removal across various scenes. The examples in Fig.~\ref{image:supp_edit} are taken from VPBench, covering diverse editing scenarios: changes in object attributes, text generation, and face identity preservation after modification. These results demonstrate VidPivot's strong performance in both video inpainting and video editing tasks. We have already provided a qualitative comparison between VidPivot and VideoPainter on the object removal task in the main text. Here, we present a comparison between VidPivot and VACE in Fig~\ref{image:supp_comp_vace}. VidPivot consistently outperforms VACE MV2V version without introducing additional artifacts, while maintaining spatial coherence and temporal consistency. 

\section{Discussion about Inference Time}
Regarding efficiency, we conducted a comparative analysis of inference time with VideoPainter and VACE, which are also capable of handling video inpainting tasks. Both VidPivot and VideoPainter are built upon the Image-to-Video (I2V) concept. The pipeline's runtime includes the generation of the entire video, incorporating the initial frame's generation time. Notably, our VidPivot pipeline also encompasses the generation of anchor frames through reference inpainting. 

When processing a 49-frame video, the VideoPainter pipeline has an inference cost of approximately 2.55 seconds per frame. In contrast, our VidPivot can process a greater number of frames, up to 81, in a single inference, with a total inference cost of about 2.59 seconds per frame, which is comparable to VideoPainter. It is important to note that VideoPainter is constrained by the limitations of its base model, with a processing cap of 49 frames per video. This necessitates more frequent segmentation of video clips and multiple rounds of inference when processing real-world video data. This, in turn, would significantly increase the total processing time. 
VACE takes approximately 260s to perform inference on an 81-frame video. This inference time is also comparable to that of VidPivot, as both methods employ a ControlNet-like architecture.
All comparative experiments were conducted using the same number and model of GPUs, and the model used for generating the initial frame is FLUX.1.Fill in all cases.

\section{User Study}
\begin{figure}
\includegraphics[width=0.95\linewidth]{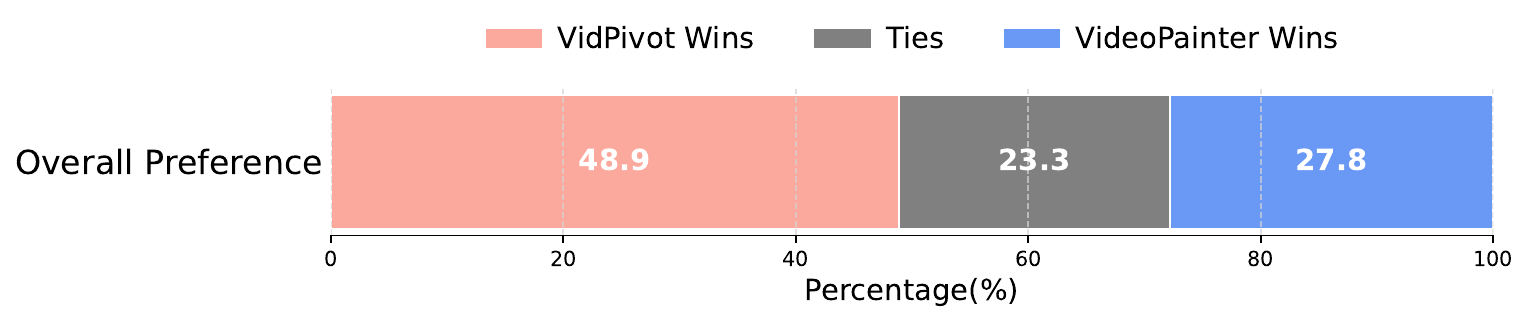}
   \caption{User study of VidPivot and VideoPainter on DAVIS benchmark.}\label{image:supp_user_study}
   \vspace{-5pt}
\end{figure}
To provide a more comprehensive evaluation of the generated video quality, we conducted a user study based on the DAVIS benchmark (includes 90 videos). We invited 20 participants for simple human evaluation. The compared methods include VideoPainter and our VidPivot, which were evaluated on overall preference, including spatial coherence and temporal consistency. The human evaluation results, presented in Fig.~\ref{image:supp_user_study}, indicate that VidPivot is rated superior to VideoPainter on the DAVIS dataset, particularly for the object removal task.

\section{Perspective and Object Dynamics}
\begin{figure}
\centering
\includegraphics[width=\linewidth]{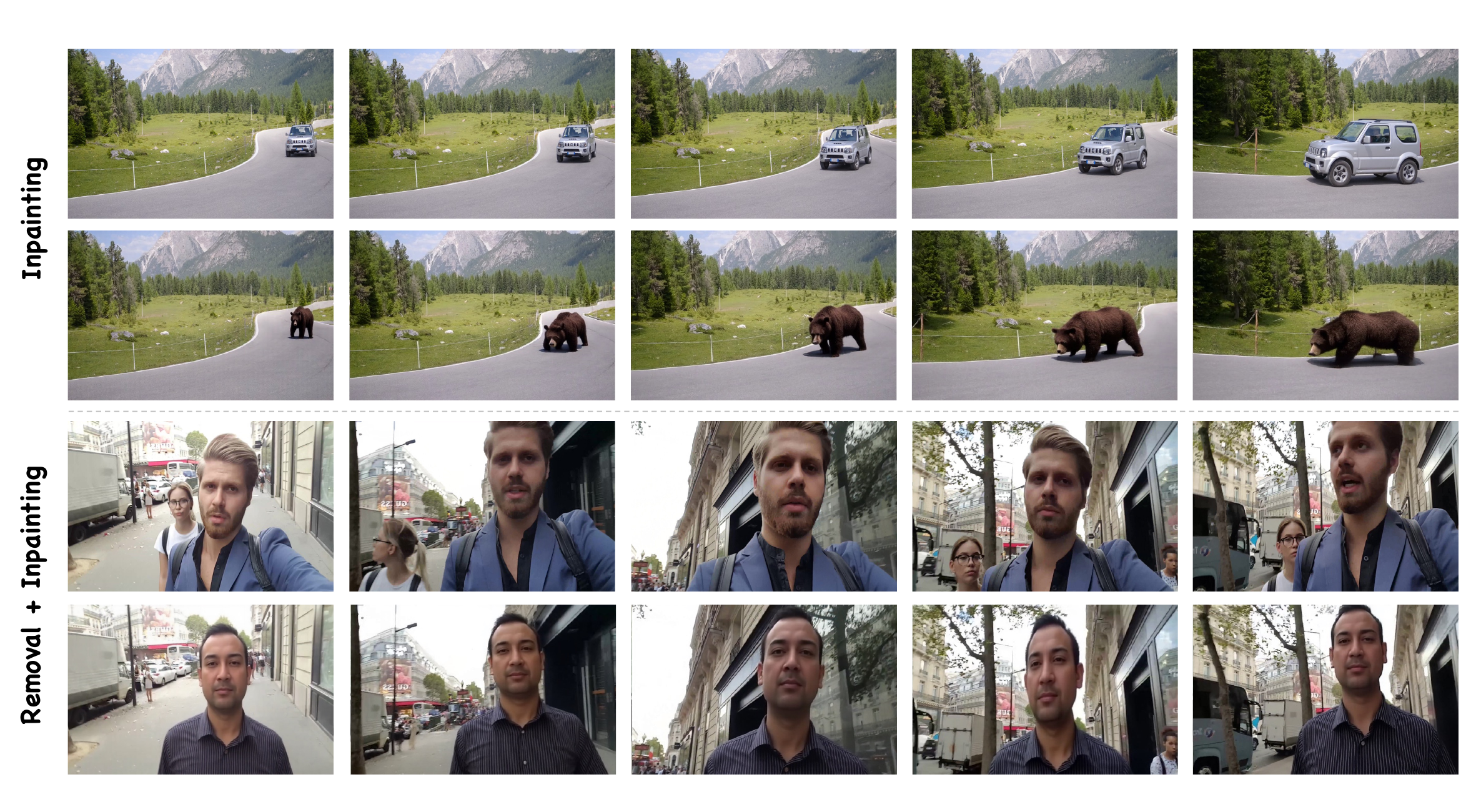}
\caption{Results from FrameProp that showcase its effectiveness in generating plausible perspectives.}\label{image:supp_perspective}
\vspace{-5pt}
\end{figure}
Addressing different perspective object dynamics is an crucial problem for video editing. The ID alignment adapter within our FrameProp module is specifically designed for this, benefiting from its training on multi-view data. As we have shown some cases, the adapter can generate objects from new viewpoints without requiring explicit conditions like camera poses. To further substantiate this capability, here we present additional intermediate results from FrameProp that showcase its effectiveness in generating plausible perspectives in Figure~\ref{image:supp_perspective}. When introducing camera pose for explicit pixel warping, reference inpainting becomes easier because the inpainting/editing region is partial fixed, but we do not consider these additional information in this work.

\end{document}